# A BAYESIAN VARIANT OF SHAFER'S COMMONALITIES FOR MODELLING UNFORESEEN EVENTS

by Robert F. Bordley *

April 16, 1993

## Abstract

Shafer's theory of belief and the Bayesian theory of probability are two alternative and mutually inconsistent approaches toward modelling uncertainty in artificial intelligence. To help reduce the conflict between these two approaches, this paper reexamines expected utility theory — from which Bayesian probability theory is derived.

Expected utility theory requires the decision maker to assign a utility to each decision conditioned on every possible event that might occur. But frequently the decision maker cannot foresee all the events that might occur, i.e., one of the possible events is the occurrence of an unforeseen event. So once we acknowledge the existence of unforeseen events, we need to develop some way of assigning utilities to decisions conditioned on unforeseen events.

The commonsensical solution to this problem is to assign similar utilities to events which are similar. Implementing this commonsensical solution is equivalent to replacing Bayesian subjective probabilities over the space of foreseen and unforeseen events by random set theory probabilities over the space of foreseen events. This leads to an expected utility principle in which normalized variants of Shafer's commonalities play the role of subjective probabilities.

Hence allowing for unforeseen events in decision analysis causes Bayesian proba-

*Operating Sciences Department; General Motors Research Labs;Warren,Michigan 48090

bility theory to become much more similar to Shaferian theory.

# 1  INTRODUCTION

Evaluating the desirability of different decisions $d$ under uncertainty requires that the decision maker

1. list every eventuality, $E \in F$, that he foresees as possible occurrences

2. add $F^c$ to this list if something unforeseen might occur. (As Fischhoff, Slovic & Lichtenstein(1978) note, many real problems involve events the decision maker did not foresee.)

3. assign each event $E$ a probability $\Pr(E)$ (with $\sum_{E \in F} \Pr(E) + \Pr(F^c) = 1$)

4. assess the utility of each decision given each foreseen event, $u(d|E)$

5. assess the utility of each decision given the unforeseen event $F^c$.

Expected utility theory then mandates choosing decision $d$ over $d^*$ if

$$\sum_{E \in F \vee F^c} \Pr(E) u(d|E) \geq \sum_{E \in F \vee F^c} \Pr(E) u(d^*|E) \qquad (1)$$

It is common to implement step 5 by assuming

**ASSUMPTION 0: For some $u_0$, $u(d|F^c) = u_0$ for all $d$.**

Given this assumption, the expected utility criterion in (1) remains valid with $\Pr(E)$ replaced by $\Pr(E|F)$, enabling the decision maker to skip steps (2) and (5).

Unfortunately Assumption 0 is often inadequate. Thus frequently the utility of decision $d$ given each atomic event $E$ can be thought of as some well-defined function of $m$ characteristics of event $E$. (See Keeney & Raiffa(1976) for further discussion of such multiattribute utility functions.) Now suppose an unforeseen event event $I_E$ occurs which

- is identical to event $E$ on almost all of these key $m$ characteristics

- is quite different from any event other than $E$ on these $m$ characteristics

Then the utility of decision $d$ given $I_E$ is likely to be much closer to the utility of decision $d$ given $E$, than it is to the utility of decision $d$ given any event other than $E$. This violates Assumption 0.

The next section formalizes this notion of similarity between events and develops a revised version of Assumption 0. This revision of Assumption 0 is equivalent to replacing subjective probability theory over the full space of events $F \vee F^c$ with normalized commonalities over the space of foreseen events $F$. Since commonalities are an alternate way of formulating the theory of lower probabilities( Good(1962), Koopman(1940), Shafer(1976)), this result suggests a new connection between subjective probability theory, random set theory(Goodman & Nguyen(1985), Kendall(1974)) and lower probabilities.

## 2 MAPPING EVENTS INTO THE POWER SET OF $F$

### 2.1 Defining Characteristics for Atomic & Compound Events

Suppose the utility of a decision given an atomic event $E$ is completely determined by the event's characteristics, $C_k(E), k = 1, ..., m$. Hence if $E$ and $E*$ have precisely the same set of characteristics (i.e., $C_k(E) = C_k(E*), k = 1...m$), then $u(d|E) = u(d|E*)$ for any $d \in D$. We define the range of each characteristic as the set of all values it assumes for the various foreseen events. Suppose we then fix all characteristics but characteristic $j$ at some reference level (e.g., their value at the status quo.) We define the importance of $j$ as how much $u(d|E)$ can vary as we vary characteristic $j$ over its range and vary $d \in D$. We then renumber the characteristics[1] in order of decreasing importance.

Consider a compound event, $A$, formed from the union of atomic events in $F$ and suppose that all atomic events in $A$ have the same first $r$ characteristics (i.e., they agree on the $r$ most important characteristics) but not the same first $r + 1$ characteristics, i.e. $C_k(E) = C_k(E^*), E, E^* \in A, k = 1...r$ but $C_{r+1}(E) \neq C_{r+1}(E^*)$ for some $E, E^* \in A$. We refer to these first $r$ characteristics as compound event A's characteristics.

---

[1] Note that the analyst could redefine his partition in terms of all possible occurrence of different $m$-tuples of characteristics. Since there are no unforeseen events in this partition, applying probability theory is straightforward. But since utilities have not been defined over all possible occurrences of various $m$-tuples of characteristics, further assumptions are needed to specify the utilities for $m$-tuples associated with unforeseen events. This need to specify utilities for unforeseen events leads to the approach used in this paper.

Suppose an unforeseen event $i$ occurs. We now relabel event $i$ as follows:

**An Algorithm for Labelling Unforeseen Events**

1. Identify how event $i$ rates on each of the $m$ key characteristics.

2. Let $r = m$.

3. Determine whether there are any compound or atomic events which have the same first $r$ characteristics as the unforeseen event.

4. Let $P$ be the union of all such events. If $P$ is nonempty, then label the unforeseen event as $I_P$.

5. If $P$ is empty and $r > 1$ then let $r = r - 1$ and return to step 3.

6. if $P$ is empty and $r = 1$, then label the unforeseen event as $I_\emptyset$.

## 2.2 Mapping $F \vee F^c$ into the power set of $F$

This algorithm maps all unforeseen events into the power set of $F$. Thus suppose our space of foreseen events consists of the three events
$$(1,1,1), (1,1,0), (0,0,1)$$
where $(1,1,0)$ indicates that event $(1,1,0)$ has a value of 1 for its most important characteristic, a value of 1 for its second most important characteristic and a value of 0 for its least important characteristic. Then if an unforeseen event $(1,0,0)$ occurs, our algorithm would relabel it as $I_{(1,1,1)\vee(1,1,0)}$.

We now similarly reclassify all atomic events. Let $A(C_1...C_m)$ be the union of all atomic events which have the same $m$ characteristics, $C_1,...C_m$. We relabel this event as $I_{A(C_1...C_m)}$. If an atomic event $E$'s characteristics are unique, then we relabel it as $I_E$.

Hence we have mapped the set of foreseen and unforeseen events, $F \vee F^c$, into the power set of $F$.

# 3   EXPECTED UTILITY WITH NORMALIZED COMMONALITIES

Consistent with Assumption 0, we assume that $u(d|I_\emptyset) = u_0, d \in D$, i.e., if an unforeseen event bearing no resemblance to any foreseen event occurs, then its utility is the same given all decisions. Since this makes the occurrence of $I_\emptyset$ irrelevant in determining the utility maximizing decision, we define new probabilities conditioned on its non-occurrence:

$$m_A = \Pr(I_A | I_\emptyset^c)$$

Then $\sum_{A \subseteq F} m_A = 1$ so that $m$ satisfies the same formal properties as Shafer's basic probability assignments. Equation(1)'s condition for preferring $d$ to $d^*$ becomes

$$\sum_{A \subseteq F} m_A u(d|I_A) > \sum_{A \subseteq F} m_A u(d^*|I_A) \qquad (2)$$

But equation (2) cannot be used to infer the optimal decision until utilities are assigned to $d, d^*$ conditioned on $I_A$.

Suppose that decision $d$ gives the same utility for all events which are subsets of some compound event $A$. This might occur if, for example, decision $d$'s utility only depended on the characteristics which all events in $A$ had in common[2]. If this were true, then we would use this utility to estimate the utility of decision $d$ given $I_A$, i.e., we would assume

**ASSUMPTION 1 :** $u(d|I_A) = u(d|E), E \in A$ **IF** $u(d|E) = u(d|E*), E, E* \in A$

Now suppose that different events in $A$ assign different utilities to decision $d$, i.e., the characteristics that all events in $A$ share in common only partially determine a decision's utility given those events. If we randomly picked an event in $A$, our best estimate of its utility for decision $d$ would be the average of the utilities which each event in $A$ assigns to $d$. We now propose to use this mean estimate to estimate a decision's utility given $I_A$, i.e., we generalize assumption[3] 1 to

**ASSUMPTION 2:** $u(d|I_A) = \frac{\sum_{E \in A} u(d|E)}{|A|}$

where $|A|$ is the number of atomic events in $A$. Given this assumption, equation (2) becomes

$$\sum_E C^N(E) u(d|E) \geq \sum_E C^N(E) u(d^*|E)$$

---

[2] There are, of course, obvious counterexamples to this presupposition. Nonetheless it is not an unreasonable rule of thumb.

[3] As the Appendix notes, one can construct alternatives to Assumption 2 which lead to much deeper linkages between subjective probabilities and lower probabilities.

458  Bordley

where

$$C^N(E) = \sum_{A \supseteq E} \frac{m(A)}{|A|}$$

are normalized commonalities (unlike Shafer's commonalities: $C(E) = \sum_{A \supseteq E} m(A)$). Thus our analysis of unforeseen events leads us to replace additive subjective probabilities by normalized commonalities.

Hence Bayesian theory, applied to a space of foreseen and unforeseen events, leads to expected utility theory with normalized commonalities replacing probabilities. If $E_1...E_m$ are the atoms of the space of foreseen events, then $\sum_{k=1}^m C^N(E_k) = 1$. But since $C^N(E_1 \vee E_2) \leq C^N(E_1)$, $C^N$ is subadditive for other partitions of the event space. To get an additive measure, we define

1. $\Pr(E) = C^N(E)$ for $E$ an atom of the space

2. $\Pr(B) = \sum_{E \in B} \Pr(E)$ for $B$ a union of several atoms

If we let $|A \& B|$ be the number of atoms common to both $A$ and $B$, then

$$\Pr(B) = \sum_A \frac{|A \& B|}{|A|} m(A)$$

Thus

$$\sum_{A \subseteq B} m(A) \leq \Pr(B) \leq \sum_{A \cap B \geq \emptyset} m(A)$$

i.e., our probability function is bounded by Shafer's upper and lower probabilities.

## 4  CONCLUSIONS

Most decision analyses ignore unforeseen events. But unforeseen events can cause a rational decisionmaker to refuse to bet on either $A$ or its complement, $F - A$, because of his concerns about the consequences of betting[4] given the unforeseen event, $F^c$. Hence Dutch book-related arguments(DeFinetti,1970; Lindley,1982) are inapplicable when unforeseen events are possible. This does not, of course, refute the Bayesian paradigm; what it does do is refute Assumption 0 which had made unforeseen events irrelevant in computing expected utility.

---

[4]Bordley & Hazen(1991) present many examples of a suspicious decision maker who, presented with a lottery that seems too good to be true, instinctively feels there must be a 'catch' somewhere.

Since Assumption 0 is refuted, an alternative way of assigning utilities to unforeseen events is necessary. This paper assumes that a decision's utility given some unforeseen event equals that decision's utility given similar foreseen events. We also specify an algorithm for classifying which set of foreseen events is 'similar' to each of the various unforeseen events.

Our procedure is equivalent to replacing subjective probabilities defined over the space of foreseen and unforeseen events by random set theory probabilities over the space of foreseen events. The subjective probabilities appearing in the standard expected utility theory formula are now replaced by normalized commonalities. So while others have viewed lower probabilities as a promising extension of Bayesian analysis(Gaines,1984; Good,1962; Smith, 1961; Williams, 1978; Dubois & Prade,1985), this paper[5] indicates that a certain variant of lower probability theory is implicit in Bayesian analysis given the appropriate assumptions on how unforeseen events are treated.

One could interpret our analysis as saying that a Bayesian decision theorist, conscious of the possibility of unforeseen events, will inflate his assessed probability for an event $E$ occurring so as to account for the occurrence of unforeseen events with consequences similar to event $E$. Since many psychological explanations of apparent irrationality in human decision making presume that individuals adjust probabilities(Hogarth & Einhorn,1990), the existence of unforeseen events may provide a partial explanation of such behavioral anomalies.

### Acknowledgements

I thank R. Jean Ruth and R. Nau for invaluable discussions of fuzzy set theory and I. Goodman for discussions of random set theory.

## References

[1] Bordley, R.F. "Fuzzy Set Theory, Observer Bias and Probability Theory." *Fuzzy Sets and Systems.* (Dec.,1989)

---

[5]consistent with earlier work(Bordley,1989) on fuzzy set theory